\documentclass[a4paper, conference]{IEEEtran}

\usepackage[english]{babel}
\usepackage[T1]{fontenc}
\usepackage[utf8]{inputenc}
\usepackage{amsmath}
\usepackage{amsthm}

\usepackage{amsfonts}
\usepackage{units}	
\usepackage{graphicx} 
\usepackage{subfigure} 
\usepackage{psfrag} 
\usepackage{iitem}

\usepackage{rotating}

\usepackage{colortbl,hhline,multirow} 
\newcolumntype{C}{>{\centering\footnotesize}m{1,2cm}}

\usepackage{moreverb}

\usepackage{listings} 
\usepackage{color}
\usepackage{cite}

\usepackage{algorithm}
\usepackage{algorithmic}
\floatname{algorithm}{Algorithm}

\begin{document}
\bstctlcite{IEEEexample:BSTcontrol}

\newcommand{\mc}[3]{\multicolumn{#1}{#2}{#3}}
\title{Convolutional Neural Networks for \\ Epileptic Seizure Prediction}

\author{\IEEEauthorblockN{Matthias Eberlein\IEEEauthorrefmark{1}, Raphael Hildebrand\IEEEauthorrefmark{1}, Ronald Tetzlaff\IEEEauthorrefmark{1}, Nico Hoffmann\IEEEauthorrefmark{2},\\ Levin Kuhlmann\IEEEauthorrefmark{3}\IEEEauthorrefmark{4}, Benjamin Brinkmann\IEEEauthorrefmark{5}, and  Jens Müller\IEEEauthorrefmark{1}}\\
\IEEEauthorblockA{\IEEEauthorrefmark{1}Technische Universität Dresden, Faculty of Electrical and Computer Engineering,\\ Institute of Circuits and Systems, 01062 Dresden, Germany\\
	Email: jens.mueller1@tu-dresden.de\\
	\IEEEauthorrefmark{2} Technische Universität Dresden, Faculty of Computer Science, \\ Computer Graphics and Visualisation, 01062 Dresden, Germany\\	
	\IEEEauthorrefmark{3} Faculty of Information Technology, Monash University, Clayton VIC 3800, Australia \\
	\IEEEauthorrefmark{4} Department of Medicine, St Vincent’s Hospital Melbourne, Fitzroy VIC 3065, Australia \\
	\IEEEauthorrefmark{5} Mayo Systems Electrophysiology Laboratory, Departments of Neurology and Biomedical Engineering, \\ Mayo Clinic, Rochester,	MN 55905, USA}
}

%


\maketitle

\begin{abstract}
Epilepsy is the most common neurological disorder and an accurate forecast of seizures would help to overcome the patient's uncertainty and helplessness. In this contribution, we present and discuss a novel methodology for the classification of intracranial electroencephalography (iEEG) for seizure prediction. Contrary to previous approaches, we categorically refrain from an extraction of hand-crafted features and use a convolutional neural network (CNN) topology instead for both the determination of suitable signal characteristics and the binary classification of preictal and interictal segments. Three different models have been evaluated on public datasets with long-term recordings from four dogs and three patients. Overall, our findings demonstrate the general applicability. In this work we discuss the strengths and limitations of our methodology.

\end{abstract} 


%
\IEEEpeerreviewmaketitle

\section{Introduction}
\label{sec:introduction}

50 million people worldwide suffer from epilepsy \cite{Ngugi2010} and for approximately 30\;\% of them the disease cannot be sufficiently controlled by medication \cite{Kwan2000}. Only 50\;\% of the patients who undergo resective surgery keep seizure free \cite{Tisi2011}. For all remaining patients the uncertainty and unpredictability of seizures belongs to the most severe disabilities \cite{Mormann2006, Schulze-Bonhage2008}. Although seizures cannot be completely prevented, a reliable forecast of their occurrence would help to overcome the helplessness of affected patients and would significantly improve their quality of life \cite{Elger2001}.

The ability to predict epileptic seizures opens completely new possibilities in neuroengineering. Subject of current research is the design and implementation of implantable closed-loop devices that continuously monitor the brain activity by electroencephalography (EEG) and analyse the recorded data in real-time. In case of an imminent seizure, the responsive device triggers an active intervention (e.g. stimulation of the vagus nerve) in order to prevent the clinical manifestation of the seizure \cite{Stacey2008}. Current neurostimulator systems are still based on the early seizure \textit{detection} \cite{Bergey2015, Geller2017}, but the therapy could dramatically improve if the device could identify abnormal activities \textit{before} seizure onset \cite{Gadhoumi2016, Nagaraj2015}.

Over the last 25 years, a lot of effort was put into the development of algorithms for the identification of changes in EEG minutes to hours before seizure onset \cite{Gadhoumi2016, Kuhlmann2018a}. However, although several proposed methods are superior to chance in statistical validations, the development of an algorithm for reliable seizure prediction with high sensitivity and specificity still remains unsolved. Up to this date, reproduction of promising results on different datasets and/or patients is one of the biggest remaining issues \cite{Mormann2016}. 

Two important stages can be observed when analysing the historical development of methods for seizure prediction. In early approaches, various groups attempted to identify precursors on the basis of single features or combinations thereof derived from the EEG time series. Numerous linear, non-linear, univariate \cite{Lehnertz1997, LeVanQuyen2008}, and multivariate measures \cite{Krug2007, Senger2016, Tetzlaff2012} were studied intensively, where a major part was motivated by the theory of non-linear dynamics \cite{Lehnertz2008, Mormann2006}. A comprehensive comparison of the different measures is given in \cite{Carney2011}. 

With the availability of new long-term iEEG recordings and the increased computing capabilities, we identified a second stage in seizure prediction, still characterised by a strict separation of \textit{feature extraction} and \textit{classification} \cite{Howbert2014}. During feature extraction, sets of various uni- and multivariate measures from time and frequency domain are derived from short segments of the time series. Subsequently, this representation of the sequences is input to one or more statistical models. Based on one part of the provided data, the so called \textit{training set}, these models are estimated for the classification of the segments. Finally, these models are evaluated on a distinct, previously unseen part of the provided data, the \textit{test set}.

In two crowd-sourced competitions (the \textit{American Epilepsy Society Seizure Prediction Challenge} \cite{Brinkmann2016} and the \textit{Melbourne-University AES-MathWorks-NIH Seizure Prediction Challenge} \cite{Kuhlmann2018}) algorithms based on random decision trees, generalised linear models (GLM), support vector machines (SVM), and convolutional neural networks (CNN) achieved a high classification performance on long-term recordings. However, a great effort of selection and optimisation of "hand-crafted" features has been performed to serve as input to typically large ensembles. Thus, these approaches are highly optimized on the specific datasets. 

In our view, the separation of feature extraction and classification is redundant and inefficient if deep neural networks (especially CNN) are utilised for binary classification, since deep learning methods have proved to find intricate structures in different levels of abstraction on highly-dimensional data \cite{LeCun2015}. CNN are widely used for EEG classification in brain-computer interfaces (BCI) \cite{Lotte2018}, the identification of epileptiform spikes from EEG \cite{Johansen2016}, automatic sleep-stage scoring \cite{Tsinalis2016}, and epileptic seizure \textit{detection} \cite{Park2018}. In seizure \textit{prediction} however, most studies proposing application of CNN are still extracting features in the form of binning and selecting spectral bands \cite{Kiral-Kornek2018, Truong2018, Brinkmann2016}.

Hence, we propose to overcome the concept of separated extraction of manual selected features and subsequent classification. By applying a CNN topology directly to the multi-channel EEG time series, appropriate representation of the data as well as suitable models for classification are directly derived from the data.  In the proposed work, different topologies are evaluated and possible benefits from including local information about electrodes are investigated. All evaluations have been made prospectively and with the use long-term data sets, containing data of different subjects that range over periods of multiple months. Therefore, the proposed results are actually reflecting performances in real-world applications.

\section{Materials and Methods}
\label{sec:materials_and_methods}

\subsection{Datasets}
\label{subsec:datasets}

\begin{table}[t]
	\renewcommand{\arraystretch}{1.3}
	\caption{Number of interictal and preictal 10-min clips of the two datasets. More information on the data characteristics are given in \cite{Brinkmann2016} and \cite{Kuhlmann2018}.}
	\label{tab:datasets_2}
	\begin{center}
		\vspace{-0.3cm}
		\begin{tabular}{ >{\centering\arraybackslash} p{1.2cm} >{\centering\arraybackslash} p{0.9cm} >{\centering\arraybackslash} p{0.9cm} >{\centering\arraybackslash} p{0.9cm} >{\centering\arraybackslash} p{0.9cm} }
			&	 \multicolumn{2}{c}{\textbf{training clips}}  & \multicolumn{2}{c}{\textbf{testing clips}}  \\
			&	 \textbf{interictal} & \textbf{preictal} & \textbf{interictal} & \textbf{preictal}  	\\		
			\hline
			\hline
			Dog 1 & 480 & 24 & 478 & 24 \\ 
			Dog 2 & 500 & 42 & 910 & 90 \\ 
			Dog 3 & 1440 & 72 & 865 & 42 \\ 
			Dog 4 & 804 & 97 & 933 & 57 \\ 
			\hline
			Patient 1 & 570 & 256 &  156 & 60  \\ 
			Patient 2 & 1836 & 222 &  942 & 60  \\ 
			Patient 3 & 1908 & 255 &  630 & 60  \\ 
			\hline
		\end{tabular}
	\end{center}
	\vspace*{-0.25cm}	
\end{table}
Our methodology was developed and evaluated on two datasets. Dataset 1 \cite{Brinkmann2016} comprises five canines with naturally occurring epilepsy and two human patients. The long-term iEEG recordings span up to one year with up to a hundred seizures. Human patients and Dog 5 were excluded from investigations in this study since their data acquisition settings differ from settings of Dogs 1-4. The included data was recorded with 16 channels and a sampling rate of \unit[400]{Hz} using the NeuroVista seizure advisory system implant \cite{Coles2013}. Therefore, bilateral pairs of four-contact strips were implanted symmetrically on each hemisphere. The data was segmented in 10-min data clips. Six of the 10-min clips extracted from \unit[66]{min} to \unit[5]{min} prior to each leading seizure are assigned as \textit{preictal}. Leading seizures are defined as seizures with a preceding seizure-free period of at least \unit[4]{h}. Interictal clips were selected similarly in groups, with randomly selected starting times a minimum of 1 week from any seizure. 

Dataset 2 \cite{Kuhlmann2018} contains iEEG data from three humans with refractory focal epilepsy that took part in a study with the NeuroVista system \cite{Cook2013}. Data was also recorded from 16 electrodes (four 4-contact strips) and sampled at \unit[400]{Hz}. The total recording time of each patient exceeds one year, the dataset contains data randomly drawn from a period of about half a year. Here, the placement of the electrode arrays was targeted to the presumed seizure focus, so no detailed information is provided in \cite{Cook2013, Kuhlmann2018} about the relative positioning of the electrode arrays. Again, only preictal data preceding leading seizures was considered following the convention of dataset 1, as described above. The data was also provided as 10-min clips, similar to dataset 1. Interictal data was randomly selected from periods with a minimum gap of \unit[3]{h} before and \unit[4]{h} after any seizure. An overview to the number of 10-min clips for training and test sets is given in Table~\ref{tab:datasets_2}.

\subsection{Models}
\label{subsec:models}

All data has been subsampled to a frequency of \unit[200]{Hz} since most of the features reported in \cite{Brinkmann2016} were calculated on frequency bands up to \unit[180]{Hz}. To improve robustness against instationarities, each channel of each 10 min-segment was individually mean centred and normalised to a standard deviation of 1. In order to capture information of frequencies as low as \unit[0.1]{Hz} which is a common high-pass cut-off frequency, all analysis have been conducted on non-overlapping segments with 3,000 samples, i.e. \unit[15]{s} in time. 

One of the main reasons for the popularity of CNN in other domains is their ability to learn suitable features, that are spatial invariant \cite{LeCun2015}. This corresponds to the capability of an algorithm to detect specific patterns regardless of the point of their occurrence. In order to exploit these advantages for the classification of EEG time series, no features have been extracted, but the clips of multi-channel time series were directly assigned to the input of a deep neural network. Therefore, the proposed networks learn to detect local patterns of samples, independent of the time and channel of their appearance. 

In this study, we investigated three different network topologies with different constraints to the implantation scheme. In each topology subsequent convolution, nonlinear activation and pooling extracts features on different time scales. As nonlinear activation, rectified linear units (ReLU) were chosen for the convolutional layers and dense layers, whereas a sigmoid output function is used in the output layer. In order to obtain classifications of the original \unit[10]{min} clips, the predictions of the corresponding \unit[15]{s} segments were averaged.

\begin{figure}[t]
	\centering
	\includegraphics[width=\columnwidth]{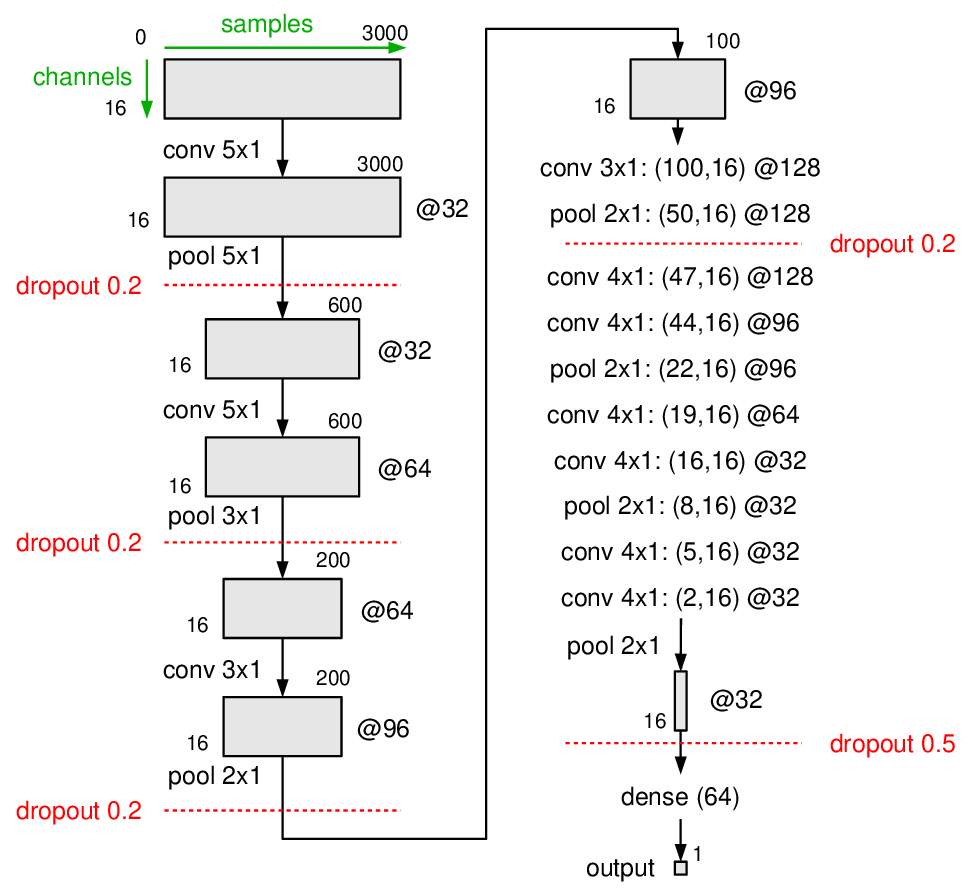}
	\caption{Illustration of the network structure of our model (topology 1). This model does not use any information about the placement of electrodes and is thus invariant to the implantation scheme. For better clarity, batch normalisation layers are not depicted.}
	\label{fig:topology1a}
\end{figure}

\subsubsection*{Topology 1} 

In topology 1 (nv1x16), convolution and pooling is performed along the time axis, which can be considered as univariate feature extraction. As depicted in Figure~\ref{fig:topology1a}, a 16-channel  segment of 3,000 samples forms the input array that is processed by a neural network with 32 layers. By subsequent convolution and pooling, the receptive field of deeper neurons grows on the time scale, i.e. the extracted features consider longer sequences of the original samples. This can be regarded as a feature extraction on lower frequency bands. Finally, the features derived from the last convolution and pooling layers are fed into a fully connected layer for classification. In order to reduce the amount of parameters in the fully connected layer, convolution and pooling has been performed until only one sample in time-axis was left. Therefore, \textit{absolute} temporal information about the occurrence of patterns is discarded but only their \textit{relative} timing is considered. In order to achieve this drastic reduction of resolution in time in an acceptable amount of convolution/pooling steps, relatively big kernel and pooling sizes of up to 5 were chosen for the first layers.

\subsubsection*{Topology 2}

As described above, only univariate features are extracted in topology 1. In order to detect local patterns that extend over multiple channels, convolution has to be performed with kernels ranging over more than just one channel. However, any algorithm extracting local patterns over multiple channels is sensitive to the arrangement of these channels in the data array. Therefore, channels were grouped according to the implantation scheme of the electrodes. In a $4\times4$ array, one dimension corresponds to the 4 electrode strips and the second dimension corresponds to the 4 contacts placed on each strip. Locality is therefore defined according to the implantation scheme. In topology 2 (nv4x4), convolution kernels stretch over multiple electrodes that are placed on one array. Therefore, this topology is able to detect patterns of samples that extend not only in time but also on multiple electrodes that are implanted next to each other on the same electrode strip. 

\subsubsection*{Topology 3}
In topology 3 (nv2x2x4), convolution and pooling is additionally performed along the axis that represents different electrode strips. The intention is to detect patterns that stretch over electrodes on different arrays. First, convolution kernels extend over electrode arrays placed in the same hemisphere, and subsequently over both hemispheres. Since the arrangement of the electrode arrays are unknown for dataset 2, this topology was only applied to subjects of dataset 1.

\subsubsection*{Training and regularisation}
All topologies have been trained and tested on each patient individually. Training data has been shuffled. ADAM \cite{Kingma2014} (learning rate of 0.001) was chosen as optimisation and binary cross entropy as loss-function. During training, class weights were used to avoid a bias on the model due to the highly imbalanced data (compare the amount of preictal to interictal segments in Table~\ref{tab:datasets_2}).
Batch normalisation \cite{Ioffe2015} was applied to the input layer and before each pooling layer. The model was additionally regularised by dropout layers with probabilities of $p=0.2$ and $p=0.5$ (see Figure~\ref{fig:topology1a}). Moreover, we used L1 and L2 regularisation, each with a factor of $10^{-9}$. In empiric investigations, informations gained by using validation turned out to be inapplicable to the test set. Therefore we refrained from using early stopping regularization and stopped training after 50 epochs.

\section{Results}
\label{sec:results}
\begin{table}[t]
	\renewcommand{\arraystretch}{1.3}
	\caption{AUC scores of the test set after patient-specific training runs for the three topologies. For dataset 2, evaluation was based on the private test set, while for dataset 1, results show the AUC of the whole test set. Every training and subsequent testing has been repeated 10 times, mean values of these 10 runs are shown. Due to unknown implantation scheme, topology 3 has not been applied to subjects of dataset 2.}
	\label{tab:auc_results}
	\begin{center}
		\vspace{-0.3cm}
		\begin{tabular}{>{\centering\arraybackslash} p{1.2cm} >{\centering\arraybackslash} p{1.4cm} >{\centering\arraybackslash} p{1.4cm} >{\centering\arraybackslash} p{1.4cm}  }
			&	 \textbf{Topology 1} & \textbf{Topology 2} & \textbf{Topology 3}	\\		
			\hline
			\hline
			Dog 1 & 0.787 & 0.769 & 0.748\\ 
			Dog 2 & 0.777 & 0.797 & 0.785\\ 
			Dog 3 & 0.825 & 0.851 & 0.802\\ 
			Dog 4 & 0.893 & 0.899 & 0.886\\ 
			\hline
			Patient 1  & 0.244 & 0.216 & - \\ 
			Patient 2 & 0.737 & 0.681 & - \\ 
			Patient 3 & 0.721 & 0.659 & - \\ 
			\hline
		\end{tabular}
	\end{center}
	\vspace*{-0.25cm}	
\end{table}

\begin{figure}[t]
	\centering
	\includegraphics[width=\columnwidth]{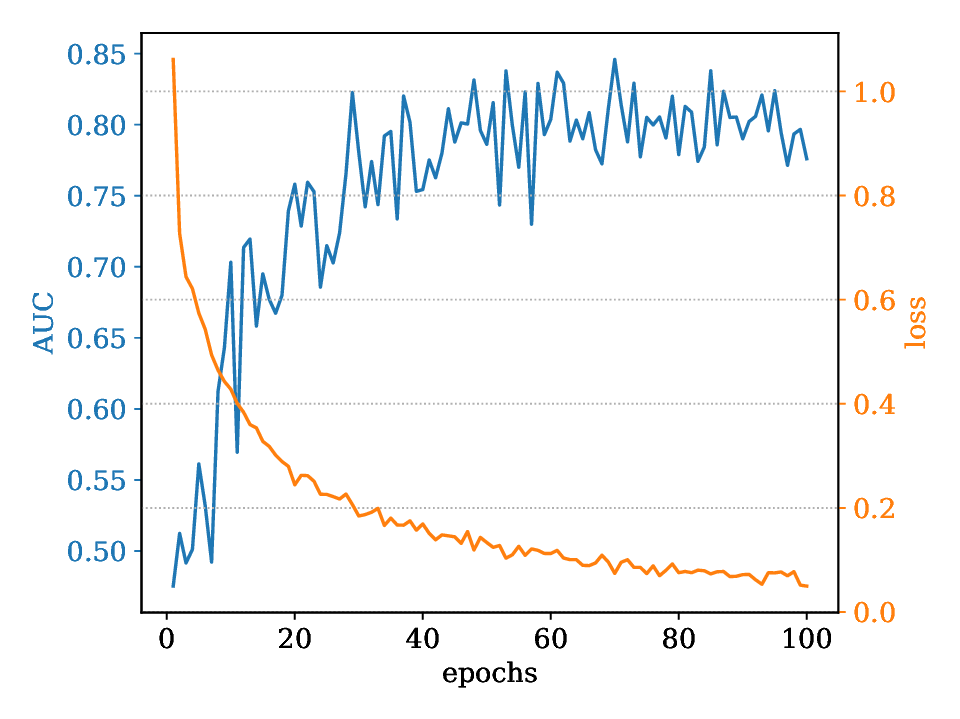}
	\caption{Evolution of the loss of the training data and AUC score of the testing data over the epochs during training (dataset 1 \textit{Dog 2})}
	\label{fig:loss_auc_over_epochs}
\end{figure}

\begin{figure*}[t]
	\centering
	\includegraphics[width=1.\textwidth]{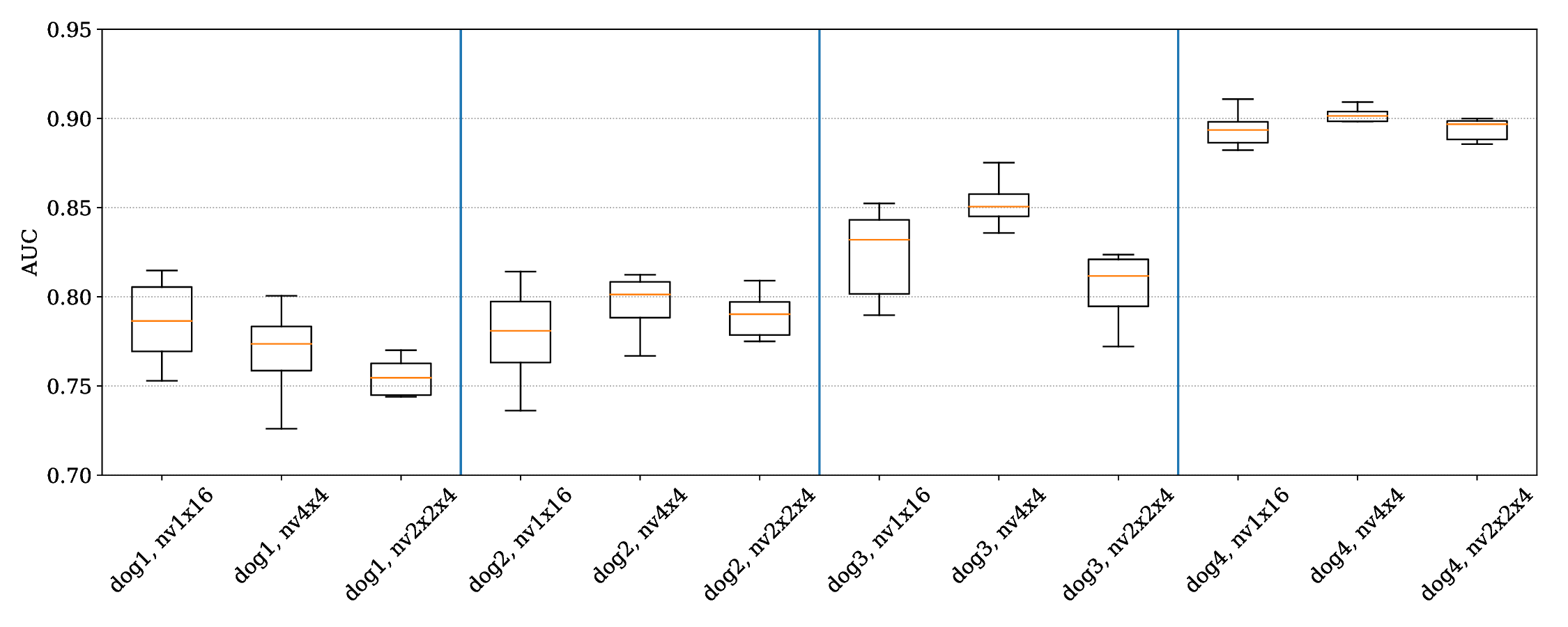}
	\caption{Box plots showing the distribution of the AUC scores of the test set for all subjects of dataset 1. A total of 10 individual runs with different initialisations have been performed for each subject of both datasets. Patients 1-3 of dataset 2 show similar variations.}
	\label{fig:boxplots1}
\end{figure*}

In \cite{Brinkmann2016, Kuhlmann2018}, classification scores were computed as area under curve (AUC) of the receiver operating characteristic (ROC). It is determined from the false positive rates and the true positive rates after applying different thresholds. An AUC score of 1.0 depicts perfectly separable predictions, while a random predictor achieves an AUC score of 0.5.   Table~\ref{tab:auc_results} presents the mean AUC scores we obtained with the three investigated topologies for all seven subjects individually. To calculate the mean AUC scores, we performed 10 independent training runs with different random seeds on each subject and each topology. It turned out that the performance is rather robust against different initialisations, as can been seen from the box plots in Figure~\ref{fig:boxplots1} (only for the dogs of dataset 1).

The training loss and the corresponding AUC score on the test set is depicted exemplarily for \textit{Dog 2} in Figure~\ref{fig:loss_auc_over_epochs}. The performance on the test set does not improve for an extended training over more epochs. It is clearly visible though that the resulting AUC score varies to some extent, depending on the exact termination of the training.

\section{Discussion}
\label{sec:discussion}

\subsection{Evaluation of the method}
\label{subsec:evaluation_of_the_method}

\begin{table}[t]
	\renewcommand{\arraystretch}{1.3}
	\caption{Comparison of the proposed method with selected studies on the same datasets. For each subject, we chose our best-performing run (from topology 2 for the dogs and from topology 1 for the patients)}
	\label{tab:comparison}
	\begin{center}
		\vspace{-0.3cm}
		\begin{tabular}{>{\centering\arraybackslash} p{1.2cm} >{\centering\arraybackslash} p{1.5cm} >{\centering\arraybackslash} p{1.7cm} >{\centering\arraybackslash} p{1.7cm}  }
			&	 \textbf{proposed method} & \textbf{winning solution\cite{Brinkmann2016}} & \textbf{winning solution\cite{Kuhlmann2018}}	\\		
			\hline
			\hline
			Dog 1 & 0.798 & 0.938 & - \\ 
			Dog 2 & 0.812 & 0.857 & - \\ 
			Dog 3 & 0.844 & 0.860 & - \\ 
			Dog 4 & 0.919 & 0.888 & - \\ 
			\hline
			Patient 1 & 0.252 & - & 0.552 \\ 
			Patient 2 & 0.751 & - & 0.735 \\ 
			Patient 3 & 0.770 & - & 0.868 \\ 
			\hline
		\end{tabular}
	\end{center}
	\vspace*{-0.25cm}	
\end{table}

We demonstrate that CNN are able to extract features on different scales from iEEG time series that allow for a separation of preictal and interictal data clips. All results are considered as a proof-of-concept of the new methodology. Except for a very recent contribution \cite{Acharya2018} on short-term datasets, the application of CNN has not been studied systematically for seizure prediction on long-term data. The hyperparameters (number of layers, number of feature maps, learning rate, size of convolution kernels, dropout probability etc. -- see Figure~\ref{fig:topology1a}) of the network have been carefully chosen with the priority of finding models with stable performance over all subjects. However, the performance can most likely be increased by an extensive optimisation and fine-tuning of these hyperparameters, which is typically a highly heuristic process. 

A comparison of our best-performing topologies to the winning solutions in \cite{Brinkmann2016} and \cite{Kuhlmann2018} is given in Table~\ref{tab:comparison}. Since we are aiming for a comparison as fair as possible, the AUC values where taken only of one of the 10 runs (the run with the highest AUC value). It has to be noted though, that the winning solutions were determined on a private leader board score that has been calculated over all testing clips independent of the subject, i.e. the AUC scores do not necessarily represent the best-performing solution of the contest for an individual subject.

To assess the generality of our models, the determination of hyperparameters has only been done by using the data (\textit{training} and \textit{testing}) of \textit{Dog 2}. Remaining subjects have been evaluated only by applying the methods that were developed on the data of \textit{Dog 2}. No further optimisation was done by analysing the performance on \textit{Dog 1}, \textit{Dog 3}, \textit{Dog 4} or any patient of dataset 2.

This puts the results given in Table~\ref{tab:auc_results} and Table~\ref{tab:comparison} into perspective: On five out of six previously unseen subjects (partly even from different datasets) the proposed models are able to predict imminent seizures with a probability significantly above chance. The exceptional behaviour of \textit{Patient 1} is not sufficiently investigated at this moment. AUC scores significantly below 0.5 imply that the model is not behaving like a random predictor but is broadly predicting the opposed class. Since this behaviour is reproducible, data of \textit{Patient 1} seems to have the exceptional property that characteristic patterns in preictal segments of the training set  occur predominantly in interictal periods in the test data or vice versa.

Based on our evaluation, we do not identify any general differences in performance of the three topologies. However, for three out of four dogs we achieve a slightly better performance with topologies that extract features over multiple channels in an early stage (see Figure~\ref{fig:boxplots1}). On the other hand, the univariate feature extraction in topology 1 shows the best performance on all three patients of dataset 2. Hence, at the moment we do not clearly see a  preference to any of our proposed models.

\subsection{Handling of datasets}
\label{subsec:handling_of_datasets}

\begin{figure}[t]
	\centering
	\includegraphics[width=\columnwidth]{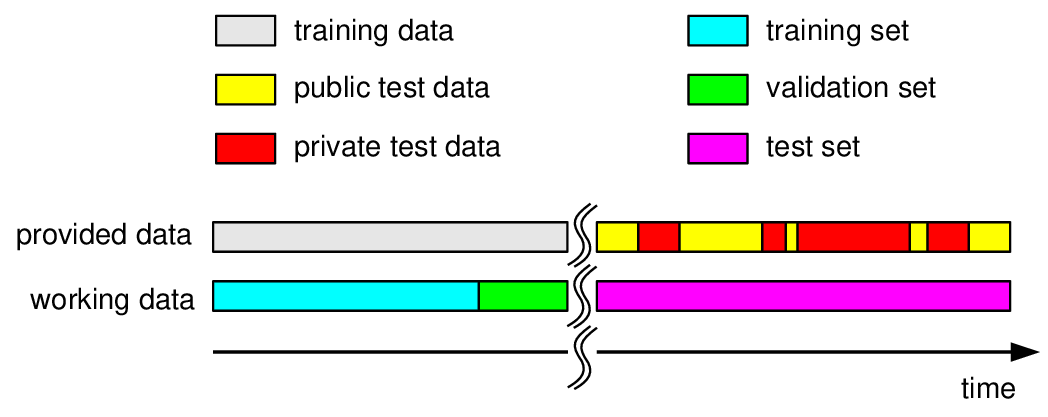}
	\caption{Partitioning of public datasets (denoted as \textit{provided data}) and its application in a machine learning scenario (\textit{working data})}
	\label{fig:data_partitioning}
\end{figure}

As already discussed in \cite{Korshunova2018}, a major problem in the comparison and benchmarking of recent methodologies is an inconsistent handling of public datasets. Typically, the datasets are separated in training data (with labels to all data clips), public test data (with labels), and finally a private data set without any information of the ground truth. The idea of the public test set is to provide data to participants of a competition for the evaluation and validation of their algorithms \cite{Brinkmann2016, Kuhlmann2018}, whereas the private test sets serve for the final scoring and are not available for the optimisation of hyperparameters.

As depicted in Figure~\ref{fig:data_partitioning}, public and private data sets are often randomly chosen data clips from a recording phase that is temporally separated from the training set. This separation is necessary to simulate a real-world scenario for a seizure prediction system, but the algorithm scoring is distorted if the public set is used for parameter tuning, since the robustness against instationarities will be increased from a retrospective point of view.

Hence, we recommend the following methodology for data handling in context of seizure prediction: If an evaluation of the trained model is used to optimise the model's hyperparameters, the \textit{validation set} should definitely be part of the provided \textit{training data}, as outlined in the bottom of Figure~\ref{fig:data_partitioning}. Especially for the training of deep neural networks, the public testing data shall not be used for regularisation by early stopping and should only be used for the evaluation of the final model \cite{Korshunova2018}. In this study, we strictly omitted any regularisation based on the public test sets. In our experimental phase, some trainings with validation set were performed by using 20\,\% of the training set for the purpose of validation. 

In order to emphasize the distortions of results that origin from different handling of the public test set, we ran another training on all three patients of dataset 2 while using the public test set for validation. If the AUC value of a model was below 0.5, its output $o$ was recalculated as $1 - o$. Again, the model was trained for 50 epochs, but instead of choosing the latest model, we used the model with the highest AUC value on the public test set. This improved the obtained model relative to the average performance of the 10 runs in Table~\ref{tab:auc_results} by 0.512, 0.044 and 0.059 for \textit{Patient 1}, \textit{Patient 2}, and \textit{Patient 3}, respectively. Since this approach implies the use of prospective data (compare Figure~\ref{fig:data_partitioning}), this does not correspond to a real test scenario as for example with an implantable device. We strongly discourage from applying such methods for improvement of results in scientific publications.

\section{Conclusion}
\label{sec:conclusion}

One clear issue is the failure of the algorithm on \textit{Patient 1} of dataset 2. The reason for this behaviour is still to be determined and conclusions have to be drawn. One possible solution could include training on data from multiple subjects to prevent overfitting on anomalies of specific datasets. In a similar way, using transfer learning techniques as e.g. pre-trained models could improve estimations by exploiting similarities between the problems of predicting seizures for different patients.

Additionally, the models themselves can certainly be improved, e.g. by using recurrent structures. The utilisation of LSTM structures might be a suitable approach to efficiently include temporal information. Furthermore, regarding the nv4x4 and nv2x2x4 topologies, neighbourhood of channels has been defined according to the implantation scheme. This does not necessarily represent the optimal model of information dependencies between channels in multichannel iEEG signals. Moreover, using ensembles of different classifiers is another possibility to improve the proposed method.

\section*{Acknowledgements}
This work was supported by the European Regional Development Fund (ERDF) and the Free State of Saxony (project number: 100320557).

We thank the Center for Information Services and High Performance Computing (ZIH) at TU Dresden for generous allocations of computer time.

\bibliographystyle{IEEEtran}
\bibliography{IEEEabrv,mlesp2018.bbl}

\end{document}